\documentclass[11pt,a4paper]{article}
\usepackage[hyperref]{emnlp2018}
\usepackage{times}
\usepackage{latexsym}

\usepackage{url}

\usepackage{amsmath}
\usepackage{graphicx} 
\usepackage{linguex}
\usepackage{multirow}
\usepackage{subcaption}
\usepackage[multiple]{footmisc}

\aclfinalcopy 

\title{Rule induction for global explanation of trained models}

\author{Madhumita Sushil, Simon \v{S}uster \and Walter Daelemans \\
  Computational Linguistics \& Psycholinguistics Research Center, \\
  University of Antwerp, Belgium \\
  {\tt firstname.lastname@uantwerpen.be} \\ }

\date{}

\begin{document}
\maketitle
\begin{abstract}
Understanding the behavior of a trained network and finding explanations for its outputs is important for improving the network's performance and generalization ability, and for ensuring trust in automated systems. Several approaches have previously been proposed to identify and visualize the most important features by analyzing a trained network. However, the relations between different features and classes are lost in most cases. We propose a technique to induce sets of if-then-else rules that capture these relations to globally explain the predictions of a network. We first calculate the importance of the features in the trained network. We then weigh the original inputs with these feature importance scores, simplify the transformed input space, and finally fit a rule induction model to explain the model predictions. We find that the output rule-sets can explain the predictions of a neural network trained for 4-class text classification from the 20 newsgroups dataset to a macro-averaged F-score of 0.80. We make the code available at \url{https://github.com/clips/interpret_with_rules}.
%
%
%
%
%
%
\end{abstract}

\section{Introduction}

Deep, non-linear neural networks are notorious for being black boxes, because the basis of a network's decision is unknown. Although sometimes we only care about better performance, understanding a trained model is important in many cases. For example, when a statistical system is used to take decisions regarding a patient's health, it is critical to know the underlying reasons.~\citet{caruana2015intelligible} have previously discussed a rule-based system that had associated the history of asthma in patients suffering from pneumonia with a lower risk of death due to it. Despite being counterintuitive, it was a predictive pattern in the data because the patients with asthma were admitted directly to the ICU and received more intensive care, which resulted in better outcomes. Model interpretability is also useful to understand the biases in the data that influence its decision. For example, explaining a trained model and its outputs can bring attention towards a potentially unfair outcome when a loan or a job opportunity is denied to an individual due to any societal bias present in the training data\footnote{\url{https://obamawhitehouse.archives.gov/sites/default/files/microsites/ostp/2016_0504_data_discrimination.pdf}}\footnote{\url{http://www.cs.toronto.edu/~madras/presentations/fairness-ml-uaig.pdf}}. Another less discussed aspect of model interpretability is its utility for analyzing a model's strengths and weaknesses. This understanding can assist with improving the model's performance and generalization ability~\citep{andrews1995survey}. 
%
%
%
%
%
%

Model interpretability techniques can either have a \textit{global} or a \textit{local} scope. A global explanation refers to the explanation of a complete model, as opposed to local explanations of individual predictions. Several existing model-agnostic interpretability techniques provide a list of important features as explanations. In such a list, the information about the interaction between different features and their correspondence to the class is lost. We propose a technique to understand the relations between the input features and the class labels that a trained supervised neural network captures. It is therefore a mechanism for global interpretability. We first weigh the input features with their importance in a trained network. We then select the best features according to the training set, and simplify them to discrete features that represent either a positive, a negative, or no correlation between a high feature value and a class label. We perform this step to limit the complexity of the output rules and make them easily understandable by humans. We  use this smaller, transformed input space to induce rules that best explain the model's predictions. We evaluate the technique on a simple text categorization problem to clearly illustrate its operation and results. We find that the output rules have a macro-averaged F-score $0.80$ when explaining the predictions of a feedforward neural network trained to classify a subset of documents from the 20 newsgroups dataset\footnote{\url{http://scikit-learn.org/stable/datasets/twenty_newsgroups.html}} into those about either `Medicine', `Space', `Cryptography', or `Electronics'.
%
%

\section{Related Work}

There has been a lot of recent interest in making machine learning models interpretable. Different approaches can be broadly grouped under two headings---1) the use of interpretable models, and 2) model-agnostic interpretability techniques. In the first case, the choice of machine learning methods is limited to the more interpretable models such as linear models and decision trees~\citep{interpretableml, caruana2015intelligible}. The drawback of incorporating model interpretability through specific model choices is that these models may not perform well enough for a given task or a given dataset. To overcome this, the second set of approaches try to explain either a complete model, or an individual prediction by using the input data and the model output(s). Several approaches involve manipulation of the trained network to identify the most significant input features. In some cases, the input features are deleted one by one, and the corresponding effect on the output is recorded~\citep{ DBLP:journals/corr/LiMJ16a, avati2017improving, DBLP:journals/corr/SureshHJCSG17}. The features that cause the maximum change in the output are ranked the highest. Another computational approach uses gradient ascent to learn the input vector that maximizes a given output in a trained network~\citep{erhan2009visualizing, simonyan2013deep}. In some other cases, the gradient of the output with respect to the input is computed, which corresponds to the effect of an infinitesimal change of the input on the output~\citep{engelbrecht1998feature, simonyan2013deep, DBLP:conf/emnlp/AubakirovaB16,SUSHIL2018103}. Another approach computes feature importance using layer-wise relevance propagation (LRP)~\citep{bach2015pixel, montavon2017explaining, Arras2017WhatIR}, which has been shown to be equivalent to the product of the gradient value and the input~\citep{kindermans2016investigating}. Sometimes the importance of a feature is analyzed by setting its value to a reference value, and then backpropagating the difference (DeepLIFT)~\citep{pmlr-v70-shrikumar17a}. In another approach, a separate `explanation model' is trained to fit the predictions of the original model~\citep{ribeiro2016should, lundberg2017unified, lakkaraju2017interpretable}. In an information theoretic approach, the mutual information between feature subsets and the model output is approximated to identify the most important features, similar to feature selection techniques~\citep{chen2018learning}. For recurrent neural networks with an attention mechanism, attention weights are often used as feature importance  scores~\citep{DBLP:conf/nips/HermannKGEKSB15, DBLP:conf/naacl/YangYDHSH16, choi2016retain}.~\citet{P18-1032} have investigated several of the previously discussed techniques and have found LRP and DeepLIFT to be the most effective approaches for explaining deep neural networks in NLP. 

Most of the above-mentioned techniques output a ranked list of the most significant features for a model. Several approaches, especially when the input is an image, visualize these features as image segments~\citep{erhan2009visualizing, simonyan2013deep, olah2018building}. These act as visual cues about the salient objects in an image for the classifier. However, such visual understanding is limited when we use either structured or textual input. Heatmaps are often used to visualize interpretations of text-based models~\citep{DBLP:conf/nips/HermannKGEKSB15, DBLP:conf/naacl/LiCHJ16, DBLP:journals/corr/LiMJ16a, DBLP:conf/naacl/YangYDHSH16, DBLP:conf/emnlp/AubakirovaB16, Arras2017WhatIR}. However, the interaction between different features and their relative contribution towards class labels remains unknown in this qualitative representation. To overcome this limitation, in the same vein as our work, rule induction for interpreting neural networks has been proposed~\citep{andrews1995survey, lakkaraju2017interpretable}.~\citet{Thrun93extractingprovably} have proposed a technique to find disjunctive rules by identifying valid intervals of input values for the correct classification. Intervals are expanded starting with the known values for instances.~\citet{lakkaraju2017interpretable} use the input data and the model predictions to learn decision sets that are optimized to jointly maximize the interpretability of the explanations and the extent to which the original model is explained. 

In our approach, we aim to generate a set of if-then-else rules that approximate the interaction between the most important features and classes for a trained model. As opposed to~\citet{lakkaraju2017interpretable}, before learning an explanation model, we modify the input data based on the importance of the features in the trained network. In doing so, we already encode some information about the network's performance within these input features.
%
%
%
%

\section{Methodology}

We are interested in identifying the if-then-else rules between different input features and class labels that are captured by a trained network to analyze the features that are the most important for classification according to the model. The insight gained in this manner can facilitate model understanding and error analysis. These rules should reflect and mimic a network's behavior for generating its output and may not correspond to human intuitions about a task, or expectations about what a network would learn. We focus on learning the rules that explain an entire model, as opposed to a single prediction. Our proposed technique comprises of these main steps:
%
%

\begin{enumerate}
\item Input saliency computation (\autoref{sec:saliency_comp})
\item Input transformation and selection (\autoref{sec:input_transform})
\item Rule induction (\autoref{sec:rule_induction})
\end{enumerate}

\begin{figure}
\includegraphics[width = \linewidth]{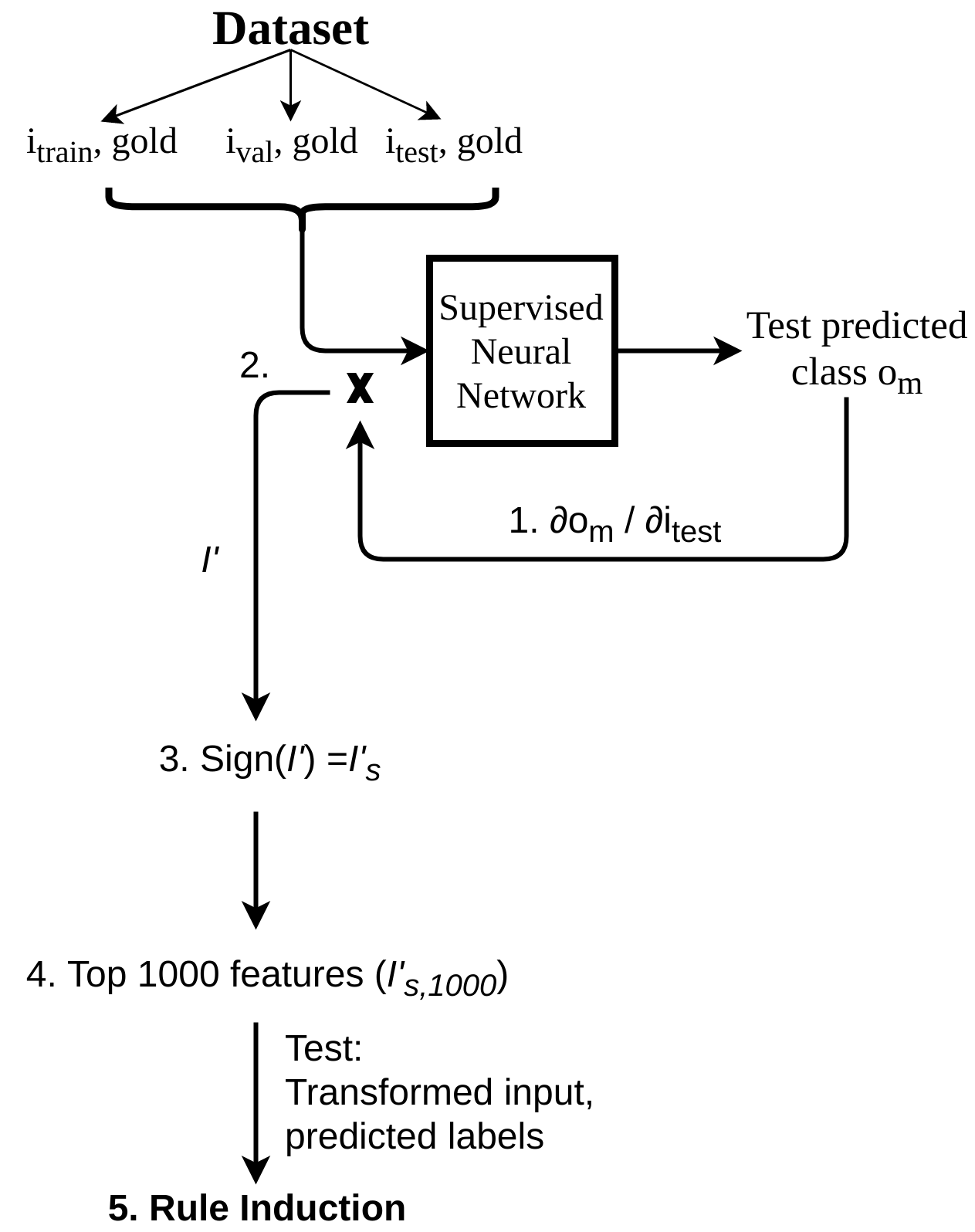}
\caption{Pipeline for rule induction for global model interpretability.}
\label{fig:pipeline}
\end{figure}

The entire pipeline is depicted in~\autoref{fig:pipeline}.

\subsection{Input saliency computation}
\label{sec:saliency_comp}

As the first step in the pipeline, we compute the contribution of the input features towards the predicted output in a trained network. This gives us the importance of different features in the network. For this, we record the change in the predicted output on modifying the input features infinitesimally; i.e., for every test instance $j$, we compute the gradient of the predicted output $o_{m}^{(j)}$ (where $m$ is the predicted output class for that instance) w.r.t.\ all the $K$ input features $i_{k}^{(j)}$, $k = 1...K$. We get a saliency map similar to~\citet{simonyan2013deep}, where the saliency $S$ of the $k$th input feature for the $j$th instance is defined as

$$ S_{k}^{(j)} = \frac{\partial o_{m}^{(j)}}{\partial i_{k}^{(j)}}. $$

\noindent Here, $m$, which is the predicted output class for that instance among all possible $n$ output classes, is computed as

$$ m^{(j)} = \underset{n}{\arg\max}(o_n^{(j)}). $$

The higher the absolute value of the gradient $S_{k}^{(j)}$, the greater the importance of the feature $k$ in the instance $j$ for the predicted class. Here, a positive sign of the saliency score indicates that the feature is positively correlated with the probability of the output class, a negative sign shows an inverse correlation, and a value of 0 shows that there is no effect of the feature on the predicted output class for that instance. 

\subsection{Input transformation and selection}
\label{sec:input_transform}

Once we have obtained the saliency scores $S$ in~\autoref{sec:saliency_comp}, we multiply these scores with the original inputs. Hence, we get transformed input data $I'$, where the input values have been reweighed according to their importance in the trained network. This corresponds to step 2 in~\autoref{fig:pipeline}.

We then reduce the transformed input data $I'$ to their sign. This is the 3rd step in the figure. This gives us a set of discrete features $I'_s$ $\in \{-1, 0, 1\} $. The value -1 indicates that the feature is highly negatively correlated with the class, i.e., a higher feature value decreases the probability of the output class. The value 1 indicates that the feature is highly positively correlated with the class, i.e., a higher value of that feature increases the probability of the output class. 0 may mean either that the feature is absent for the document, or that it is not important for the output class\footnote{We have two interpretations for 0 because we take a product of the gradients and the input feature values. 0 value of either of these two terms could transform the final value to 0.}. We perform this sign reduction step because the rule conditions with these discrete feature values are more interpretable and readable than those containing continuous reweighed vector values.

We then keep only the top $1000$ features in the trained network, represented by step 4 in the figure. We restrict the feature space to reduce the complexity of the rule induction step. We use either an unsupervised technique---sensitivity analysis, or the mutual information between the inputs $I'_s$ and the corresponding training labels. The first technique uses only the gradient values to find the most important features, whereas the second supervised technique makes use of both the transformed inputs $I'_s$ and the labels for this purpose.

\begin{itemize}

\item \textbf{Sensitivity analysis}

For feature selection using sensitivity analysis~\citep{engelbrecht1998feature}, we first compute the gradients of all the output nodes with respect to all the input features for all the instances. We then aggregate these gradient values across the instances by taking a root mean square value. The squaring ensures that negative and positive effects of a feature are treated in an equivalent manner. Hence, we obtain the overall importance scores of all the features for every output node in the network. Now, we use the maximum importance of the features across all the output nodes as the significance of the features in the trained network. The features with the highest significance scores are then selected as the top features. Hence, the method uses only the trained network weights and the original input data for feature selection. It does not make use of the labels for the instances and is hence an unsupervised technique for selecting the most important features in a trained neural network.
 
\item \textbf{Mutual information}

In this step, we identify the top features using mutual information between the reweighed features $I'_s$ (computed as the product of gradients and the original inputs, and then reduced to the corresponding sign), and the labels in the training data.

%
%

\end{itemize}

%
%

\subsection{Rule induction}
\label{sec:rule_induction}

We train a rule induction model on the transformed features $I'_s$ obtained as a result of the previous step (\autoref{sec:input_transform}) to fit the output predictions of the original model. For multi-class problems, we induce the rules in a one-vs-rest manner where the rules for explaining every individual class are found one at a time. This gives us separate discriminatory rules for all the classes. Separate rule-sets for individual classes are more interpretable compared to an ordered set of rules for multiple classes at once. In the latter case, we often need to take into account the rules that have been first learned for other classes to interpret the rules for the class we are interested in, which increases its complexity, especially when we have a large number of classes.

\subsubsection*{RIPPER-k}

We use the implementation of the rule induction algorithm RIPPER-k~\citep{cohen1995fast} in Weka~\citep{hall2009weka} (JRIP). The algorithm generates a set of if-then-else rule by first overfitting the conditions on a \textit{growing set}, and then pruning them based on their performance on a \textit{pruning set}. These rules are learned in  one-vs-rest manner in  order of increasing class prevalence, where the final else condition covers the majority class. 

In the growing phase, starting with an empty set, the algorithm adds conditions that test the values of discrete and continuous features in the dataset to attribute them to the corresponding class. For example, given two input features $f_1$ and $f_2$, the algorithm checks if the concerned class is covered by the rules $f_1=d_1$, $f_2 \leq c_1$ or $f_2 \geq c_1$, where $d_1$ is a valid value of the nominal feature $f_1$, and $c_1$ is the value of a continuous feature $f_2$ that occurs in the training data. The conditions are added repeatedly to maximize an information gain criterion. Next, the final sequence of the obtained conditions are removed one at a time to increase the generalization of the rule on the pruning set. When deleting conditions does not improve the error rate any more, pruning is terminated. Thereby, the instances covered by the rule are removed, and the process is repeated for the rest of the instances, until more than half of the instances covered by a rule in the pruning data are incorrect.
%
%

\section{Experimental Details}

\subsection{Data}

We use the documents related to `Space', `Medicine', `Electronics' and `Cryptography' from the 20-newsgroups dataset for text classification. We limit ourselves to 4 classes under the `Science' category to reduce experimental complexity. There are approximately 535 training instances, 60 development instances, and 395 test instances for every category. The development set is used for optimizing the model we want to explain. We featurize the data as a bag-of-words with TF-IDF values after removing headers, signature blocks, quotation blocks and stopwords. We get 30,346 input features in this manner.

\subsection{Model to be explained}

We explain a feedforward neural network that has been trained for 4-class text classification. The neural network has 2 hidden layers with 100 units each, ReLU activation function for these layers, and a softmax output layer. It has been optimized using the Adam optimizer~\citep{kingma2014adam} for 50 epochs to minimize the cross entropy loss. We get a macro-averaged F-score of 0.82 on the test set. 
%
%

\subsection{Metrics}
\textit{Fidelity} refers to the extent to which an interpretability technique explains the original model. It can be expressed using many different metrics. We quantify fidelity as the macro-averaged F-score of predicting the output of the model that is being explained using the rules that are induced by the explanation technique. The F-scores for the individual classes obtained in the one-vs-rest manner are averaged to compute this overall fidelity.

\subsection{Hyperparameter optimization}
\label{sec:hyperparam_opt}

We found that the rules induced using RIPPER-k are sensitive to its hyperparameters, especially to the minimum number of correctly covered instances and the seed value chosen for randomizing the instances, particularly when the dataset is small. To account for the variation, we run RIPPER-k with 50 different seed values, and the value of the minimum number of instances positively covered by a rule ranging between 2 and the number of instances of the class being explained. For each run, we compute the macro-averaged F-score for explaining the predictions of the neural network. In doing so, we obtain a standard deviation of around 10\%, 18\%, 17\% and 14\% for the classes `Space', `Medicine', `Electronics', and `Cryptography' respectively. This shows that it is important to find an optimum performance over several runs. 

We select the rule-set that results in the maximum score, and hence is the one that explains the original model predictions the best among the possible alternatives. We select the rule-set with the maximum score instead of the most generalizable RIPPER-k model because we are not interested in transferring the rules to unknown tasks. If we do not ensure that the learned rules approximate the patterns in the original model to the best possible extent, it remains unclear whether an unintuitive rule-set is obtained because of the parameters of RIPPER-k, or because our neural network has poor explanations. We compare different rule-sets with high F-scores to verify their consistency, which has been discussed in~\autoref{sec:rule_consistency}.

\section{Results and Discussion}

\subsection{Rules as explanations}

\begin{table*}
\begin{center}
\begin{tabular}{|l|rrr|rrr|}
\hline 

\multirow{2}{*}{\textbf{Text class}} & \multicolumn{3}{c|}{\textbf{MI}}  & \multicolumn{3}{c|}{\textbf{SA}}  \\
			 & \bf P & \bf R & \bf F & \bf P & \bf R & \bf F \\ \hline
Space        & 0.99 & 0.82 & 0.90 & 0.99 & 0.76 & 0.86 \\
Medicine     & 0.94 & 0.68 & 0.79 & 0.92 & 0.70 & 0.79 \\
Electronics  & 0.89 & 0.64 & 0.75 & 0.92 & 0.63 & 0.75 \\
Cryptography & 0.97 & 0.61 & 0.75 & 0.99 & 0.61 & 0.75 \\
\hline
\bf Macro-average & 0.95 & 0.69 & 0.80 & 0.96 & 0.68 & 0.79 \\
\hline
\end{tabular}
\end{center}
\caption{\label{tab:fidelity} Precision (P), Recall (R), and F-score (F) when explaining neural network predictions using the induced rules with features selected using mutual information (MI) and sensitivity analysis (SA).}
\end{table*}

We obtain a fidelity score of 0.80 using the proposed technique when the features are pre-selected using the mutual information (MI) score between the transformed inputs and the output labels. Hence, the learned set of if-then-else rules can explain the output of our neural network for 4-class text classification to an F-score of 0.80. The precision, recall, and F-scores for individual classes is presented in~\autoref{tab:fidelity}. The precision of the rules is high, which shows that the rules, if induced, are reliable. The largest F-score of 0.90 is obtained for the class \textit{space}, and the lowest F-score of 0.75 is obtained for the classes \textit{electronics} and \textit{cryptography}. On analyzing the rule-sets obtained for different classes, we see that the complexity of the rules for the \textit{space} class is lower than that of \textit{electronics}. For \textit{space} classification, single terms are often indicative of the correct class. On the other hand, for \textit{electronics} and \textit{cryptography} classes, the rules often consist of multiple words, which are jointly used to discriminate between different classes. This suggests that \textit{cryptography} and \textit{electronics} are more confusable classes, which is also reflected in their lower F-scores.

The results with MI feature selection are comparable to those obtained using sensitivity analysis feature selection, presented in~\autoref{tab:fidelity} as well. The observed patterns in the corresponding rule sets are also very similar. Hence, we only present the rules induced using MI feature selection, and those using sensitivity analysis can be found in the Appendix.

\begin{figure*}[h!]
  \fbox{\begin{minipage}{\textwidth}
  
if \textit{(just = -1) and (use = 1) $\implies$ electronics} $(24/24)$ \\
elif \textit{(circuit = 1) $\implies$ electronics} $(32/32)$ \\
elif \textit{(just = -1) and (don = 1) $\implies$ electronics} $(11/11)$ \\
elif \textit{(people = 0) and (used = 1) and (key = 0) and (don = 0) and (use = 0) and (edu = 0) and (medication = 0) and (concept = 0) and (did = 0) $\implies$ electronics} $(36/43)$ \\
elif \textit{(electronics = 1) $\implies$ electronics} $(17/18)$ \\
elif \textit{(battery = 1) $\implies$ electronics} $(23/23)$ \\
elif \textit{(radio = 1) and (shack = 1) $\implies$ electronics} $(9/9)$ \\
elif \textit{(people = 0) and (thanks = 1) and (advance = 1) $\implies$ electronics} $(12/14)$ \\
elif \textit{(signal = 1) $\implies$ electronics} $(13/15)$ \\
elif \textit{(people = 0) and (company = 1) and (just = 0) $\implies$ electronics} $(13/18)$ \\
elif \textit{(pc = 1) $\implies$ electronics} $(16/19)$ \\
elif \textit{(people = 0) and (use = 1) and (just = 0) and (good = 0) and (clipper = 0) and (probably = 0) and (center = 0) and (unless = 0) and (18084tm = 0) and (algorithms = 0) $\implies$ electronics} $(29/33)$ \\
elif \textit{(appreciated = 1) and (time = 0) $\implies$ electronics} $(11/16)$ \\
elif \textit{(voltage = 1)} $\implies$ electronics $(8/8)$ \\
elif \textit{(program = -1)} $\implies$ electronics $(10/15)$ \\
else: others (1134/1281)
  \end{minipage}}
  \caption{Set of if-then-else rules that explain the predictions of the neural network for the `electronics' class when using mutual information feature selection. Here the discrete test value 1 means a positive correlation between a feature value (which we can approximate to relative frequency due to the use of TF-IDF vectors) and the probability of the class, -1 means a negative correlation, and 0 shows an absence of a feature. The values $(a/b)$ mean that $a$ of $b$ instances covered by the rule are correct. The rules with lower values of $a/b$ are less trustworthy, and the rules with lower value of $b$, especially in the first few conditions, are less generalized.}
%
%
  \label{fig:rules_electronics_test}
\end{figure*}

Now, in~\autoref{fig:rules_electronics_test}, we present the rules that explain the predictions of the neural network for the \textit{electronics} class. The coverage of the individual rules in these sets is also reported in the format $a/b$, which means that the rule covers $b$ instances in the dataset, out of which $a$ instances are correctly covered. A higher value of $b$ suggests that a rule is more generalized, especially if it is higher up in the hierarchy where it has more instances of the correct class to its disposal. The value of $a/b$ should be used to assess how trustworthy a rule is, with lower values indicating less trustworthiness. On inspecting the gradients, we found that the gradient value was 0 only for one feature for one instance in the test set. This feature is not present in the induced rules. Hence, in these rules, the feature value of 0 has only one interpretation---the absence of the feature. The rules for the other classes can be found in the Appendix. 

There are several rules which associate class-specific content words with the corresponding class. In complex rules, we find several terms that are used to identify the \textit{electronics} class by excluding the likelihood of the other classes. For example, in the rule:

\ex. \textit{(people = 0) and (used = 1) and (key = 0) and (don = 0) and (use = 0) and (edu = 0) and (medication = 0) and (concept = 0) and (did = 0) $\implies$ electronics}

the absence of the term \textit{medication} is used to rule out the possibility of the \textit{medicine} class, the absence of the term \textit{key} is used to rule out the possibility of the class \textit{cryptography}. Hence, the model uses an elimination strategy in combination with class-relevant features for its predictions.

One of the rules learned for the class \textit{space} is:

\ex. \textit{(idea = 1) and (probably = -1) $\implies$ space}

This rule, which is matched after some other rules\footnote{The complete rule set can be found in the Appendix.} with higher coverage, covers 3 correct and 0 incorrect instances. The value of -1 associated with \textit{probably} in conjunction with 1 for \textit{idea} shows that a low score of probably, combined with a high score of idea corresponds to the class space. Similarly, in~\autoref{fig:rules_electronics_test}, we often see that the presence of the word \textit{just} reduces the probability of \textit{electronics} class. Similar associations can be observed in the rule:

\ex. \textit{(don = -1) and (just = -1) $\implies$ medicine}

where the presence of \textit{don't}\footnote{\textit{don't} is tokenized as \textit{don} and \textit{'t}, and the mention of \textit{don} is found in the rule.} and \textit{just} reduces the probability of the \textit{medicine} class. This rule covers 4 positive and 0 negative examples. These rules show that function words, which may hint towards terms related to modality, are often important for the network to classify the class of the text documents. 

\begin{figure}[h!]
\includegraphics[width=\linewidth]{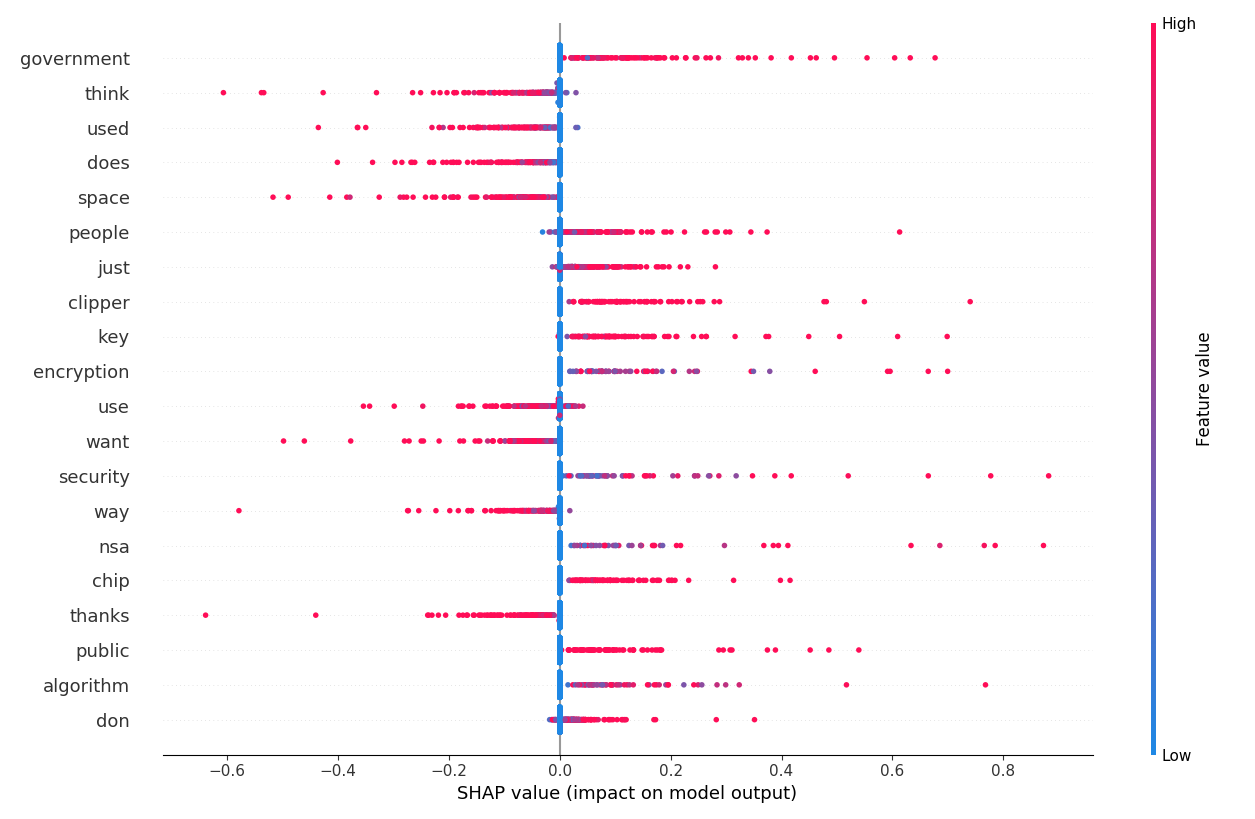}
\caption{Feature importance estimation using SHAP for model interpretability.}
\label{fig:shap}
\end{figure}

In~\autoref{fig:shap}, we present the most important features for the same model, as generated by SHAP~\citep{lundberg2017unified}. This is a state-of-the-art tool that unifies several popular approaches for model interpretability. The top features presented here overlap with the features we identify using our technique. However, the interaction between these features is not obvious from the figure. It is possible to generate an interaction plot in SHAP, where the interaction between different features is visualized. However, only a few features can be compared against each other in this manner without making the process expensive. These interactions are not understandable without extensive analysis. The technique we propose instead manages to capture the associations between features and classes, alleviating the need for these complex visualizations to understand the model.

\begin{figure*}[t!]

\begin{subfigure}{\linewidth}
  \fbox{\begin{minipage}{\textwidth}

if \textit{(circuit = 1) $\implies$ electronics} (53/55) \\
elif \textit{(people = 0) and (electronics = 1) $\implies$ electronics} (31/34) \\
elif \textit{(people = 0) and (power = 1) and (time = 0) and (research = 0) and (years = 0) $\implies$ electronics} (43/55) \\
elif \textit{(people = 0) and (thanks = 1) and (used = 1) and (particular = 0) $\implies$ electronics} (12/14) \\
elif \textit{(people = 0) and (space = 0) and (voltage = 1) $\implies$ electronics} (22/22) \\
elif \textit{(people = 0) and (motorola = 1) $\implies$ electronics} (16/19) \\
elif \textit{(people = 0) and (space = 0) and (line = 1) and (encryption = 0) and (med = 0) and (doesn = 0) and (case = 0) $\implies$ electronics} (28/39) \\
elif \textit{(people = 0) and (space = 0) and (wire = 1) $\implies$ electronics} (14/18) \\
elif \textit{(space = 0) and (people = 0) and (1174 = 1) $\implies$ electronics} (9/9) \\
elif \textit{(amp = 1) $\implies$ electronics} (13/14) \\
elif \textit{(just = 0) and (space = 0) and (8051 = 1) $\implies$ electronics} (8/8) \\
else: others (1565/1848)

  \end{minipage}}
  \caption{The rules induced on the \textbf{original training data} that point towards the associations in the data that we would expect the neural network to capture.}
  \label{fig:rules_electronics_train_original}
  \end{subfigure}
  
  \begin{subfigure}{\linewidth}
  
    \fbox{\begin{minipage}{\textwidth}
if \textit{(circuit = 1) $\implies$ electronics} $(53/53)$ \\
elif \textit{(people = 0) and (power = 1) and (time = 0) and (national = 0) $\implies$ electronics} $(48/63)$ \\
elif \textit{(people = 0) and (electronics = 1)} $\implies$ electronics $(27/28)$ \\
elif \textit{(people = 0) and (voltage = 1) $\implies$ electronics} $(22/23)$ \\
elif \textit{(people = 0) and (space = 0) and (line = 1) and (want = 0) and (government = 0) and (block = 0) and (years = 0) and (amateur = 0) and (cell = 0) $\implies$ electronics} $(28/33)$ \\
elif \textit{(people = 0) and (space = 0) and (advance = 1) $\implies$ electronics} $(18/26)$ \\
elif \textit{(people = 0) and (motorola = 1) $\implies$ electronics} $(16/16)$ \\
elif \textit{(people = 0) and (space = 0) and (wire = 1) and (digital = 0) $\implies$ electronics} $(14/14)$ \\
elif \textit{(think = 0) and (buy = 1) and (government = 0) $\implies$ electronics} $(14/21)$ \\
elif \textit{(people = 0) and (space = 0) and (uucp = 1) $\implies$ electronics} $(10/12)$ \\
elif \textit{(space = 0) and (government = 0) and (amp = 1) $\implies$ electronics} $(13/15)$ \\
elif \textit{(people = 0) and (space = 0) and (8051 = 1) $\implies$ electronics} $(8/8)$ \\
else: others $(1562/1823)$
  \end{minipage}}
  \caption{The rules induced on the \textbf{training data transformed according to the network weights}. These point towards the associations that the network actually captures, as opposed to those we expect it to capture. To find these rules, we compute the gradients of the output of the corresponding gold label class w.r.t.\ the input features, instead of taking the gradients of the output predictions.}
  \label{fig:rules_electronics_train_transformed}
  \end{subfigure}
  
  \caption{The rules induced on the training data to fit the gold labels.}
\end{figure*}

Next, in~\autoref{fig:rules_electronics_train_original}, we present the if-then-else rules that have been induced from the training data using the original input and the gold labels, to give an idea about the relations between features and classes that we would expect the model to pick up from the data. In~\autoref{fig:rules_electronics_train_transformed}, we compare them with the feature-class associations that are instead captured by the trained network. These have been identified by using the transformed input space for rule induction, also to explain gold labels. Although the rules in these ordered sets are not directly comparable, we see that there are three exactly matching rules in the two sets (ignoring the order) and several common feature conditions. In these cases, the patterns in the training data are approximated by the network. We find that the rules fit on the network-transformed training data have a 2\% higher macro averaged F-score compared to those on the original data. This suggests that the generalization brought about by the network assists in rule induction for this dataset on these tasks.

\subsection{Consistency of the induced rule-sets}
\label{sec:rule_consistency}

As we discussed in~\autoref{sec:hyperparam_opt}, it is important to optimize the hyperparameters of RIPPER-k to find an optimum set of rules because we obtain high standard deviations of the fidelity scores of explanations across different hyperparameters. Additionally, if several sets of explanations have high fidelity scores, they should also be consistent with each other. To this end, we analyze multiple rule-sets with high fidelity scores to check whether they are similar to each other. For this comparison, we identify all the sets of rules whose F-scores lie within 1\% of the F-score of the best model. However, a comparison between different sets of ordered rules is not trivial. We calculate the mean percentage of the exactly matching rules between the well-performing models and the final selected set of rules, when these sets are assumed to be unordered. This score penalizes the rules that match partially as being a mismatch, which makes it a strict metric for sets with longer rules. In this process, we get the scores in the range 50\%--79\% using MI, and 48\%--90\% using sensitivity analysis feature selection techniques. We find that the classes with high fidelity scores also have high rule overlap, and vice versa. This suggests that different models that explain less confusable classes are also more consistent across different parameters. We additionally calculate the mean percentage of the instances that have been classified identically by the well-performing models and the final selected model to facilitate a semantic comparison. We find that the classification overlap between different models ranges from 94\%--97\% when using MI feature selection, and 93\%--98\% with sensitivity analysis feature selection. The exact numbers can be found in the Appendix.

\section{Limitations}

While the advantage of using a rule inducer like RIPPER-k lies in gaining insight into feature-class associations, the approach has its own drawbacks. RIPPER-k outputs rules according to class prevalence. Hence, the majority class only has an `else' clause associated with it. Furthermore, only the default rule is fired when there is just one class in the dataset. Hence, this technique is unsuitable for one-class problems, and when the class we are interested in is the majority class in the one-vs-rest binary setup. Moreover, if several features frequently co-occur and infrequently occur without each other, this technique may find only one of them.
%
%

Next, the proposed technique is a global explanation technique that can be used to identify the if-then-else rules that explain a model as a whole. However, using this technique, we can not obtain such rules for explaining only a single instance.

Finally, the learned rules are sensitive to some parameters in RIPPER-k. As discussed earlier, we overcome this limitation by optimizing the performance over different parameters. However, this step reduces the speed of finding explanations.

\section{Conclusions and Future work}

In this paper, we have proposed a technique to learn if-then-else rules to explain the predictions of supervised models. We have first computed the gradients of the output predictions with respect to the input features for every instance. We have then rescaled these gradients to feature weights, and have multiplied them with the original inputs to learn reweighed inputs for every instance. We have then simplified them to a set of 1000 transformed features with discrete values. Finally, we have induced rules that combine these features into rule conditions for every class in the data separately. We have found that the induced rules can explain the predictions of our classifier to a macro-averaged F-score of 0.80. We have shown that these rules can be used to understand a model's behavior and output predictions.

In future, we plan to evaluate the proposed technique on different datasets to compare the fidelity scores of the explanations across datasets with different complexities. We would also like to compare our work with other techniques for inducing rules as explanations. It would also be interesting to investigate other rule induction algorithms that can support one-class problems, and that are less sensitive to parameters such as shuffling of data to overcome the limitations present due to the use of RIPPER-k.
%
%

\section*{Acknowledgments}
We would like to thank the anonymous reviewers for their useful comments. This research was carried out within the Accumulate strategic basic research project, funded by the government agency Flanders Innovation \& Entrepreneurship (VLAIO) [grant number 150056]. 

\bibliographystyle{acl_natbib_nourl}
\bibliography{emnlp2018}

\appendix

\section{Appendix}
\label{sec:supplemental}

In~\autoref{tab:rule_consistency}, we present the mean percentage of the exact match between different rule-sets output by several good models obtained with different RIPPER-k parameters, compared to the rule-set we finally selected. In the same table, we also present the mean percentage of instances that have been classified by different models identically as the model we finally selected. Furthermore, in~\autoref{fig:rules_sens_test}, we present the rules induced after performing feature selection using sensitivity analysis to explain the predictions of the neural network on the test data. In~\autoref{fig:rules_mi_test}, we present the rules induced when we instead use mutual information for feature selection to explain the same predictions. 

\begin{table*}
\begin{center}
\begin{tabular}{|l|rr|rr|}
\hline 

\multirow{2}{*}{\textbf{Text class}} & \multicolumn{2}{c|}{\textbf{MI}}  & \multicolumn{2}{c|}{\textbf{SA}}  \\
			 & \bf Rule match & \bf Classification match & \bf Rule match & \bf Classification match \\ \hline
Space        & 79\% & 97\% & 90\% & 98\% \\
Medicine     & 63\% & 96\% & 68\% & 94\%\\
Electronics  & 52\% & 94\% & 57\% & 93\% \\
Cryptography & 50\% & 97\% & 48\% & 97\% \\
\hline
\end{tabular}
\end{center}
\caption{\label{tab:rule_consistency} Mean exact match between unordered set of rules from several good models obtained with different parameters of RIPPER-k and the selected model for every class, and the mean classification overlap between them.}
\end{table*}

\begin{figure*}

\begin{subfigure}{\textwidth}
  \fbox{\begin{minipage}{\textwidth}
  
if  \textit{(just = -1) and (use = 1) $\implies$ electronics} (24/24) \\
elif \textit{(just = -1) and (like = 1) $\implies$ electronics} (14/14) \\
elif \textit{(circuit = 1) $\implies$ electronics} (32/32) \\
elif \textit{(electronics = 1) $\implies$ electronics} (24/25) \\
elif \textit{(battery = 1) $\implies$ electronics} (22/22) \\
elif \textit{(people = 0) and (used = 1) and (way = 0) and (clipper = 0) and (space = 0) and (good = 0) and (fairly = 0) and (drug = 0) $\implies$ electronics} (38/48) \\
elif \textit{(line = 1) and (space = 0) and (encryption = 0) and (clipper = 0) and elif (medical = 0) and (doctor = 0) $\implies$ electronics} (20/20) \\
elif \textit{(people = 0) and (thanks = 1) and (advance = 1) and (long = 0) $\implies$ electronics} (10/10) \\
elif \textit{(people = 0) and (voltage = 1) $\implies$ electronics} (10/11) \\
elif \textit{(company = 1) and (medical = 0) and (order = 0) and (minutes = 0) and elif (clipper = 0) $\implies$ electronics} (19/23) \\
elif \textit{(pc = 1) and (security = 0) $\implies$ electronics} (12/12) \\
elif \textit{(think = -1) and (didn = -1) $\implies$ electronics} (5/5) \\
elif \textit{(people = 0) and (cheap = 1) $\implies$ electronics} (10/14) \\
elif \textit{(just = 0) and (motorola = 1) $\implies$ electronics} (6/6) \\
elif \textit{(just = 0) and (tape = 1) $\implies$ electronics} (6/7) \\
elif \textit{(end = -1) $\implies$ electronics} (7/10) \\
else: others (1144/1296)
  \end{minipage}}
  \caption{Rules for the electronics class}
  \label{fig:rules_sens_test_electronics}
  \end{subfigure}
  
  \end{figure*}
  
  \begin{figure*}[t!] \ContinuedFloat
  
  \begin{subfigure}{\linewidth}
  \fbox{\begin{minipage}{\textwidth}
  
if \textit{(medical = 1) $\implies$ medicine} (55/55) \\
elif \textit{(doctor = 1) $\implies$ medicine} (38/38) \\
elif \textit{(cause = 1) and (time = 0) $\implies$ medicine} (30/35) \\
elif \textit{(disease = 1) $\implies$ medicine} (17/17) \\
elif \textit{(body = 1) $\implies$ medicine} (25/31) \\
elif \textit{(med = 1) $\implies$ medicine} (14/14) \\
elif \textit{(effects = 1) and (space = 0) $\implies$ medicine} (17/21) \\
elif \textit{(like = -1) and (time = 1) $\implies$ medicine} (7/7) \\
elif \textit{(don = 0) and (skin = 1) $\implies$ medicine} (7/7) \\
elif \textit{(photography = 1) $\implies$ medicine} (13/13) \\
elif \textit{(cancer = 1) $\implies$ medicine} (8/8) \\
elif \textit{(surgery = 1) $\implies$ medicine} (8/8) \\
elif \textit{(pain = 1) $\implies$ medicine} (6/8) \\
elif \textit{(allergic = 1) $\implies$ medicine} (7/7) \\
elif \textit{(water = 1) and (make = 0) $\implies$ medicine} (9/13) \\
elif \textit{(left = 1) and (use = 0) $\implies$ medicine} (7/9) \\
elif \textit{(don = 0) and (blood = 1) $\implies$ medicine} (5/5) \\
elif \textit{(don = 0) and (experience = 1) $\implies$ medicine} (7/9) \\
elif \textit{(therapy = 1) $\implies$ medicine} (4/4) \\
else: others (1147/1270)
  \end{minipage}}
  \caption{Rules for the medicine class}
  \label{fig:rules_sens_test_med}
  \end{subfigure}
  
  \end{figure*}
  
  \begin{figure*}[t!] \ContinuedFloat
  
  \begin{subfigure}{\linewidth}
  \fbox{\begin{minipage}{\textwidth}
  
if \textit{(space = 1) $\implies$ space} (116/118) \\
elif \textit{(orbit = 1) $\implies$ space} (29/29) \\
elif \textit{(earth = 1) $\implies$ space} (23/24) \\
elif \textit{(sky = 1) $\implies$ space} (17/17) \\
elif \textit{(nasa = 1) $\implies$ space} (12/12) \\
elif \textit{(launch = 1) $\implies$ space} (14/14) \\
elif \textit{(solar = 1) $\implies$ space} (9/9) \\
elif \textit{(moon = 1) $\implies$ space} (7/7) \\
elif \textit{(shuttle = 1) $\implies$ space} (6/6) \\
elif \textit{(spacecraft = 1) $\implies$ space} (6/6) \\
elif \textit{(ground = -1) and (secret = 0) $\implies$ space} (4/4) \\
elif \textit{(plane = 1) $\implies$ space} (3/3) \\
elif \textit{(materials = 1) and (st = 0) $\implies$ space} (4/4) \\
else: others (1245/1326)
  \end{minipage}}
  \caption{Rules for the space class}
  \label{fig:rules_sens_test_space}
  \end{subfigure}
  \end{figure*}
  
  \begin{figure*} \ContinuedFloat
  
  \begin{subfigure}{\linewidth}
  \fbox{\begin{minipage}{\textwidth}
  
if \textit{(clipper = 1) and (moon = 0) $\implies$ cryptography} (90/90) \\
elif \textit{(key = 1) and (care = 0) and (like = 0) $\implies$ cryptography} (37/38) \\
elif \textit{(government = 1) and (launch = 0) and (medical = 0) and (nasa = 0) $\implies$ cryptography} (45/46) \\
elif \textit{(encryption = 1) $\implies$ cryptography} (25/25) \\
elif \textit{(nsa = 1) $\implies$ cryptography} (13/13) \\
elif \textit{(just = 0) and (david = 1) and (want = 0) and (disease = 0) $\implies$ cryptography} (13/15) \\
elif \textit{(time = 0) and (algorithm = 1) $\implies$ cryptography} (8/8) \\
elif \textit{(com = 1) and (metzger = 1) $\implies$ cryptography} (9/9) \\
elif \textit{(good = 0) and (modem = 1) $\implies$ cryptography} (6/6) \\
elif \textit{(does = 0) and (crypto = 1) $\implies$ cryptography} (8/8) \\
elif \textit{(just = 0) and (don = 0) and (security = 1) $\implies$ cryptography} (7/7) \\
else: others (1145/1314)
  \end{minipage}}
  \caption{Rules for the cryptography class}
  \label{fig:rules_sens_test_cryptography}
  \end{subfigure}
  
  \caption{Set of if-then-else rules that \textbf{explain the test predictions} of the neural network for the all the classes when \textbf{using sensitivity analysis for feature selection}. Here the discrete test value 1 means a positive correlation between a feature value (which we can approximate to relative frequency due to the use of TF-IDF vectors) and the probability of the class, -1 means a negative correlation, and 0 shows the absence of a feature. The values $(a/b)$ mean that $a$ of $b$ instances covered by the rule are correct. The rules with lower values of $a/b$ are less trustworthy, and the rules with lower value of $b$, especially in the first few conditions, are less generalized.}
  \label{fig:rules_sens_test}
\end{figure*}

\begin{figure*}

\begin{subfigure}{\linewidth}
  \fbox{\begin{minipage}{\textwidth}
  
if \textit{(just = -1) and (use = 1) $\implies$ electronics} $(24/24)$ \\
elif \textit{(circuit = 1) $\implies$ electronics} $(32/32)$ \\
elif \textit{(just = -1) and (don = 1) $\implies$ electronics} $(11/11)$ \\
elif \textit{(people = 0) and (used = 1) and (key = 0) and (don = 0) and (use = 0) and (edu = 0) and (medication = 0) and (concept = 0) and (did = 0) $\implies$ electronics} $(36/43)$ \\
elif \textit{(electronics = 1) $\implies$ electronics} $(17/18)$ \\
elif \textit{(battery = 1) $\implies$ electronics} $(23/23)$ \\
elif \textit{(radio = 1) and (shack = 1) $\implies$ electronics} $(9/9)$ \\
elif \textit{(people = 0) and (thanks = 1) and (advance = 1) $\implies$ electronics} $(12/14)$ \\
elif \textit{(signal = 1) $\implies$ electronics} $(13/15)$ \\
elif \textit{(people = 0) and (company = 1) and (just = 0) $\implies$ electronics} $(13/18)$ \\
elif \textit{(pc = 1) $\implies$ electronics} $(16/19)$ \\
elif \textit{(people = 0) and (use = 1) and (just = 0) and (good = 0) and (clipper = 0) and (probably = 0) and (center = 0) and (unless = 0) and (18084tm = 0) and (algorithms = 0) $\implies$ electronics} $(29/33)$ \\
elif \textit{(appreciated = 1) and (time = 0) $\implies$ electronics} $(11/16)$ \\
elif \textit{(voltage = 1)} $\implies$ electronics $(8/8)$ \\
elif \textit{(program = -1)} $\implies$ electronics $(10/15)$ \\
else: others (1134/1281)
  \end{minipage}}
  \caption{Rules for the electronics class}
  \label{fig:rules_mi_test_electronics}
  \end{subfigure}
  \end{figure*}
  
  \begin{figure*} \ContinuedFloat
  \begin{subfigure}{\linewidth}
  \fbox{\begin{minipage}{\textwidth}
  
if \textit{(medical = 1) $\implies$ medicine} (55/55) \\
elif \textit{(doctor = 1) $\implies$ medicine} (38/38) \\
elif \textit{(body = 1) $\implies$ medicine} (28/34) \\
elif \textit{(effects = 1) and (don = 0) and (earth = 0) $\implies$ medicine}(18/19) \\
elif \textit{(disease = 1) $\implies$ medicine} (19/19) \\
elif \textit{(photography = 1) $\implies$ medicine} (13/13) \\
elif \textit{(med = 1) $\implies$ medicine} (15/15) \\
elif \textit{(cause = 1) and (station = 0) and (enforcement = 0) and (antennas = 0) and (attacks = 0) and (battery = 0) $\implies$ medicine} (24/26) \\
elif \textit{(allergic = 1) $\implies$ medicine} (7/7) \\
elif \textit{(experience = 1) and (data = 0) $\implies$ medicine} (9/13) \\
elif \textit{(surgery = 1) $\implies$ medicine} (10/10) \\
elif \textit{(skin = 1) $\implies$ medicine} (7/7) \\
elif \textit{(blood = 1) $\implies$ medicine} (6/7) \\
elif \textit{(pain = 1) $\implies$ medicine} (6/8) \\
elif \textit{(therapy = 1) $\implies$ medicine} (4/4) \\
elif \textit{(cancer = 1) $\implies$ medicine} (5/5) \\
elif \textit{(food = 1) $\implies$ medicine} (4/4) \\
elif \textit{(just = -1) and (don = -1) $\implies$ medicine} (4/4) \\
elif \textit{(cells = 1) $\implies$ medicine} (5/7) \\
else: others (1154/1284)

  \end{minipage}}
  \caption{Rules for the medicine class}
  \label{fig:rules_mi_test_med}
  \end{subfigure}
  
  \end{figure*}
  
  \begin{figure*} \ContinuedFloat
  
  \begin{subfigure}{\linewidth}
  \fbox{\begin{minipage}{\textwidth}
  
if \textit{(space = 1) $\implies$ space} (116/118) \\
elif \textit{(orbit = 1) $\implies$ space} (29/29) \\
elif \textit{(earth = 1) $\implies$ space} (23/24) \\
elif \textit{(sky = 1) $\implies$ space} (17/17) \\
elif \textit{(nasa = 1) $\implies$ space} (12/12) \\
elif \textit{(launch = 1) $\implies$ space} (14/14) \\
elif \textit{(moon = 1) $\implies$ space} (7/7) \\
elif \textit{(solar = 1) $\implies$ space} (9/9) \\
elif \textit{(shuttle = 1) $\implies$ space} (6/6) \\
elif \textit{(spacecraft = 1) $\implies$ space} (6/6) \\
elif \textit{(atmosphere = 1) $\implies$ space} (4/4) \\
elif \textit{(idea = 1) and (probably = -1) $\implies$ space} (3/3) \\
elif \textit{(18084tm = 1) $\implies$ space} (4/4) \\
elif \textit{(gamma = 1) $\implies$ space} (3/3) \\
elif \textit{(exploration = 1) $\implies$ space} (3/3) \\
elif \textit{(landing = 1) $\implies$ space} (3/3) \\
elif \textit{(aircraft = 1) $\implies$ space} (3/3) \\
elif \textit{(ground = -1) and (accepted = 0) $\implies$ space} (4/4) \\
elif \textit{(materials = 1) and (aids = 0) $\implies$ space} (3/3) \\
elif \textit{(rocket = 1) $\implies$ space} (2/2) \\
else: others (1245/1305)
  \end{minipage}}
  \caption{Rules for the space class}
  \label{fig:rules_mi_test_space}
  \end{subfigure}
  \end{figure*}
  
  \begin{figure*} \ContinuedFloat
  \begin{subfigure}{\linewidth}
  \fbox{\begin{minipage}{\textwidth}
  
if \textit{(clipper = 1) and (delta = 0) $\implies$ cryptography} (90/90) \\
elif \textit{(key = 1) and (care = 0) $\implies$ cryptography} (49/54) \\
elif \textit{(government = 1) and (money = 0) and (develop = 0) $\implies$ cryptography} (37/39) \\
elif \textit{(encryption = 1) $\implies$ cryptography} (23/23) \\
elif \textit{(nsa = 1) $\implies$ cryptography} (14/14) \\
elif \textit{(com = 1) and (metzger = 1) $\implies$ cryptography} (9/9) \\
elif \textit{(crypto = 1) $\implies$ cryptography} (11/11) \\
elif \textit{(time = 0) and (algorithm = 1) $\implies$ cryptography} (8/8) \\
elif \textit{(used = 0) and (court = 1) $\implies$ cryptography} (6/6) \\
elif \textit{(security = 1) $\implies$ cryptography} (8/9) \\
elif \textit{(used = 0) and (modem = 1) $\implies$ cryptography} (5/5) \\
else: others (1141/1311)
  \end{minipage}}
  \caption{Rules for the cryptography class}
  \label{fig:rules_mi_test_cryptography}
  \end{subfigure}
  
  \caption{Set of if-then-else rules that \textbf{explain the test predictions} of the neural network for the all the classes when \textbf{using mutual information for feature selection}. Here the discrete test value 1 means a positive correlation between a feature value (which we can approximate to relative frequency due to the use of TF-IDF vectors) and the probability of the class, -1 means a negative correlation, and 0 shows the absence of a feature. The values $(a/b)$ mean that $a$ of $b$ instances covered by the rule are correct. The rules with lower values of $a/b$ are less trustworthy, and the rules with lower value of $b$, especially in the first few conditions, are less generalized.}
  \label{fig:rules_mi_test}

\end{figure*}

\end{document}